\documentclass[12pt,letterpaper]{article}

\usepackage[letterpaper,margin=1in]{geometry}
\usepackage{setspace}

\usepackage{graphicx}
\usepackage{subcaption}
\usepackage{float}
\usepackage{rotating}
\usepackage{booktabs}
\usepackage{multirow}
\usepackage{pdflscape}

\usepackage{amsmath}
\usepackage{amssymb}
\usepackage{amsfonts}

\usepackage[numbers,square,sort&compress]{natbib}
\usepackage{xcolor}
\usepackage{textcomp}
\usepackage{url}
\usepackage[colorlinks=true,citecolor=blue,linkcolor=blue,urlcolor=blue]{hyperref}
\hypersetup{pdfstartview=FitH}

\newcommand{\SafeIncludeGraphics}[2][]{%
  \IfFileExists{#2}{\includegraphics[#1]{#2}}{%
    \begingroup\setlength{\fboxsep}{5pt}%
    \fbox{\parbox[c][30mm][c]{0.82\linewidth}{\centering\small
    Figure file not uploaded:\par\texttt{\detokenize{#2}}}}%
    \endgroup}}

\title{A Hybrid Visitor Prediction Framework using Interpretable Temporal features, Integrating IQR and Anomaly Detection in Lakes in Nebraska}

\author{
\textbf{Israt Zarin}\\
School of Computing\\
University of Nebraska--Lincoln\\
Lincoln, Nebraska 68588, USA\\
\texttt{izarin2@huskers.unl.edu}
\and
\textbf{Christopher Chizinski}\\
School of Natural Resources\\
University of Nebraska--Lincoln\\
Lincoln, Nebraska 68583, USA\\
\texttt{cchizinski2@unl.edu}
}

\date{}

\begin{document}

\maketitle

\begin{abstract}
For effective resource management,  infrastructure planning, and sustainable tourism, it is really important to predict accurate visitor numbers. This paper estimates the daily visitation of Nebraska lakes using spatial, basic temporal, and interpretable temporal features. Moreover, we evaluate the base models with multiple metrics. Using the best three models, our method will ensemble hybrid models and choose the best hybrid technique. Further, we investigate the impact of lag features and how the models work with different outlier detection techniques such as  Interquartile Range(IQR) and Anomaly detection. The models are evaluated using  regression metrics (MAE, MSE, RMSE, MAPE,and $R^2$). Experiment result shows that, after deducted anomaly, it reduces  MAE, MSE,RMSE,and MAPE. After adding lag, it is showing some positive effect on the result. Moreover, Anomaly identification techniques is more better than IQR for improving performance and preserve data. The reported anomaly-based cross-validation results favored the voting ensemble over the compared base models on the evaluated metrics. In anomaly-based cross-validation, VOTE improved prediction errors compared with MLP (256,128,64); improvements varied in the single-split comparison. Specifically, the hybrid model reduced MAE by 9.82\%, MSE by 3.12\%, RMSE by 1.76\%, and MAPE by 6.77\%. In the reported cross-validation comparison, the best anomaly-based hybrid model had lower MAE, MSE, RMSE, and MAPE than the best IQR-based hybrid model by 9.81\%, 34.30\%, 18.99\%, and 0.66\%, respectively.
\end{abstract}


\section{Introduction}\label{sec1}

The management of recreation areas, parks, museums, and shielded landscapes is becoming more dependent on time and space-rich data on when and where individuals visit\cite{cite27}\cite{cite28}. Old techniques, including manual counters, intercept survey methods and periodic survey methods, have been useful in giving snapshots but do not meet the management demands of the scale, frequency or geographic location distribution of current visitation due to the labour-intensive nature and expensive tools of these methodologies\cite{cite33}, their high cost when used at high frequencies, their spatially limited nature and their inability to capture short-term fluctuations in behaviour and behavioural shifts across multiple sites at once. 

The mobile/GPS mobility data has reached maturity over the last 10 years and has made it possible to track human movement continuously and on a fine scale, so it is now possible to observe the daily, weekly, and seasonal patterns of human movement over a large number of sites at the same time \cite{cite28}.

Now a days, machine learning techniques are used to estimate visitors. With the development of the field, there has been a variety of research in terms of forecasting visitors\cite{cite1}\cite{cite2} \cite{cite31} and being effective in predicting planning, waste management, safety, and conservation planning. Cultural institutions and destinations: Studies show that tree-based models (e.g., Gradient Boosting, Random Forest) and sequence models can be even better than classical baselines by taking advantage of calendar structure (weekday, holidays, season) and exogenous drivers (weather, events). To illustrate, visit the forecasting at museums. It has found recurring patterns, and special-day effects are better described using ensembles than using linear baselines \cite{cite1}, whereas studies of lakes and parks have found that decision trees and decision-related classifiers can be very successful at classifying the expected demand \cite{cite2}. Similar to city-level demand prediction, in urban areas, hybrid deep learning, namely, LSTM with gradient boosting, has demonstrated better results in city-level predictions as compared to the ARIMA-like models, indicating the worth of both nonlinear features interaction and short-term memory \cite{cite3}\cite{cite30}. Syntheses of the tourism-ML literature also record the trend to data-driven, hybrid, and ensemble methods with more power to their content and stacked learners to elevate accuracy and stability \cite{cite4}. An emerging body of work on the problem of sparse geolocation and regional modeling demonstrates how the addition of spatial structure (i.e. graph-based relationships between places) can improve the level of prediction where direct observations are incomplete or heterogeneous across locations \cite{cite5}\cite{cite6}

This study is needed due to the fact that erroneous or noisy mobility information may result in erroneous information on how often visitors are visited and thereby alter funding and staffing, as well as policy decisions in managing recreational resources. As an example, raw spikes or drops in mobility indicators can either overestimate or underestimate the true visitation patterns, thereby giving false direction on conservation planning or investing in infrastructure \cite{cite34}. By effectively reducing the problem through interquartile range (IQR), strong anomaly detection, the accuracy of machine learning models and the credibility of the decisions based on the data are enhanced. Not only that, but previous studies seldom test the behaviour of ensemble approaches after the removal of anomalies, so it is unclear which factor is responsible for greater improvement in performance: data cleaning or the type of model.

Meanwhile, scholars have also grown sensitive to the data-quality facts of mobility streams. Raw signals may consist of duplicated pings, GPS drift, device/platform peculiarities, and transmission gaps, which may be passed up into modelling pipelines. According to empirical studies, transport and smartphone traces may contain systematic noise and platform differences, which, when not mitigated, will result in biased downstream analytics and poor performance \cite{cite7}\cite{cite8}. Expansive reviews of the spatio-temporal mining emphasise that robust pipelines should incorporate quality management and anomaly handling as opposed to more advanced models, particularly when resources are allocated, and policies are based on outputs\cite{cite9}. Simultaneously, the spatio-temporal analytics society has suggested multi-scale anomaly detection on human-activity data and quality screening on volunteered geographic information, showing that the detection and pre-filtering of abnormal observations can significantly enhance the trustworthiness of the following model \cite{cite10}\cite{cite11}.

Although there is an evident improvement, there are four gaps. To begin with, the anomaly handling is not very often explicit. The vast majority of tourism-forecasting pipelines do not explicitly detect outliers, either by pre-aggregating or unintentionally cleaning data points, however outliers can overstate or understate estimated visitation and behave to reinforce the effects of downstream features\cite{cite17}. Second, these comparisons of models are frequently limited: most studies compare a single family (e.g., tree ensembles or a particular deep network) and do not compare the variability of different regressors with similar splits and metrics, which makes it difficult to separate the benefits of data and the benefits of the model choice. Moreover, a single model can be biased. Third, the use of interpretable features: despite recent models that focus on it, most visitor-count analyses continue to work with flat tabular features and cannot be generalised between sites and seasons, and they don't focus much on interpretable features\cite{cite5}\cite{cite8}. Lastly, when the models are showing close results, how to identify which one is best? 

The research proposes using extra temporal features and lag features, an interquartile range (IQR), and an anomaly-aware system; it cleaned dataset and  predicted visitor numbers from extensive mobility datasets. The research design consists of two distinct phases. The first stage employs three unsupervised anomaly detection methods, including Isolation Forest, Local Outlier Factor, and One-Class SVM, to remove abnormal data points while preserving genuine visitation patterns \cite{cite15}\cite{cite16}\cite{cite17}. Moreover, at the same time, applying IQR to remove outliers helped to find out which process is more effective.  The second stage performs identical model evaluations between base model candidates like tree-based models, linear and regularised models, and neural networks  \cite{cite12}\cite{cite13}\cite{cite14}, which receive the same data partitions and evaluation metrics. The research design compares model performance across anomaly removal and IQR filtering, while also considering differences among model types. The research applies existing evidence from museums and lakes, marine parks, and urban tourism to create daily visitation estimates for recreation management decisions \cite{cite1}\cite{cite2} \cite{cite3}\cite{cite4}\cite{cite5}\cite{cite6} and also examined them with hybrid methods. It also used multiple metrics for evaluation so that it would be easy to choose between models whose performances are very similar.
This study examines five factors relative to the accuracy of visitor prediction and the strength of model evaluation. First, it investigated the possibility of better prediction  by using anomaly detection and the Interquartile Range (IQR) method without compromising data quality and investigated which method was more effective. Second, it evaluated whether hybrid machine learning models can outperform individual machine learning models. Third, it examines whether we need a single metric to assess models or whether a full multi-metric framework is needed for good assessment. Fourth, it examined  the role of interpretable temporal features and lag variables as additional factors to increase the accuracy of predictions. Finally, if several models have similar predictive capabilities, it determines which model offers the best possible predictions in terms of accuracy, reliability and robustness. It develops a hybrid method that helps agencies convert noisy mobility data into reliable analytical tools for their staffing, conservation, and infrastructure planning needs.

\section{Materials and Methods}
The study presents an anomaly-conscious machine learning model to predict visitor activity at the recreational lakes within Nebraska based on large-scale mobility data. Its workflow consists of five main stages that include data acquisition and preprocessing, feature engineering, anomaly detection, model development and evaluation, and visualisation.

\begin{figure}[H]
    \centering
    \SafeIncludeGraphics[width=0.9\textwidth]{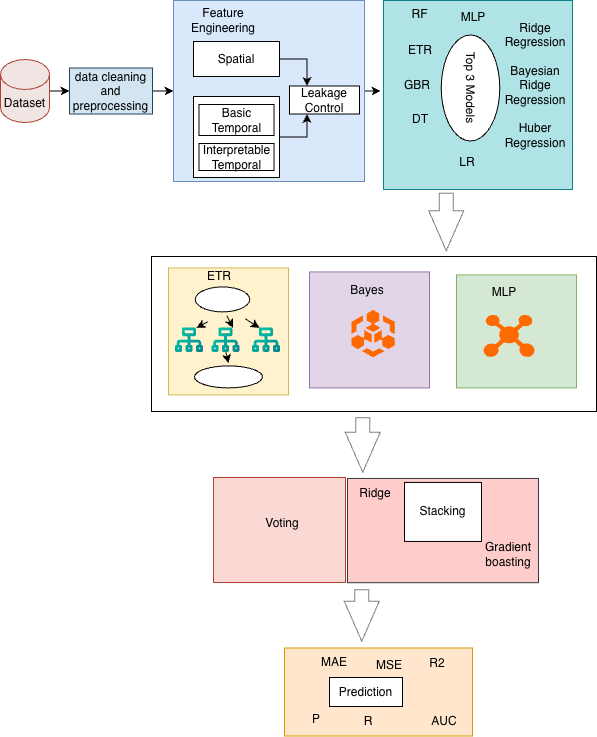}
    \caption{Overview of the lake-visitation prediction workflow, including preprocessing, feature engineering, anomaly handling, base-model evaluation, and hybrid-model development.}
    \label{fig:workflow}
\end{figure}

The workflow presented in diagram \ref{fig:workflow}  represents the end-to-end modelling pipeline that is used in this research. We begin with the summarised visitation data and then proceed to carry out data cleaning and preprocessing to achieve consistency and reliability. This is followed by a well-structured aspect of feature engineering, in which spatial features, pure temporal features, and interpretable temporal features (like calendar and hourly visit patterns) are generated, and explicit leakage control is implemented to avoid using information that may unreasonably overpower predictive power. An ensemble of single regression models is then trained and evaluated (and can include tree-based, linear, and regularised models and neural networks), and the best-performing models are picked. The chosen models then act as base learners in a hybrid modelling phase, where they are both voted on and stacked, and ridge regression and gradient boosting are also used as meta-learners so that the individual models can be strengthened and weaknesses can be complementary to each other. Lastly, a wide range of regression- and classification-based measures, such as MSE, MAE, R2, precision, recall, and AUC, are used to evaluate model performance in terms of an integrated measure of predictive accuracy and predictive strength in various data regimes.

\subsection{Data Collection and Cleaning}
The data will be gathered through the process of data collection and preprocessing. The data employed in the present research were obtained based on the AirSage mobility platform that combines anonymous and geographical visitation data throughout the waterbody in the state of Nebraska. The records are the estimated number of visitors to a site in a day and other metrics, including the time spent and the time information. In this project, we used 123627 records of 227 lakes that were gathered and processed. The data had blanks, duplicate records, and inconsistencies of identifiers that necessitated massive cleaning. The preprocessing procedures were:
\subsection{Dataset preprocessing}
In this research, we used a cleaned lake-visitation dataset where each row contains a lake identifier (\texttt{poi\_id}), date (YYYYMMDD), and target visitation count. Missing data are removed from the dataset. Features are divided into two categories. 
We used a cleaned, daily lake-visitation dataset where each record contains a lake identifier (\texttt{poi\_id}), a date (date, parsed from YYYYMMDD), and the target variable count (daily visitation). Records with missing data  were removed. If the dataset contained an alternative lake identifier column (e.g., \texttt{POI\_ID}, \texttt{lake\_id}, \texttt{id}), it was renamed to \texttt{poi\_id}. To provide a consistent numeric representation of lake identity, we created \texttt{poi\_id\_code} using categorical encoding of \texttt{poi\_id}.
\begin{itemize}
    \item Change of date attributes to a standard format \texttt{YYYY\-MM\-DD}.
    \item Assuring a unitary numeric representation of visitor numbers and visit durations.
    \item Elimination of zero-value records or incomplete entries.
    \item Arranging the records in order of \texttt{poi\_id} (point of interest identifier) and date to maintain temporal order.
\end{itemize}

It was used to clean data, meaning that it provided data uniformity and reliability before training the models.

\subsection{Feature Engineering}
At first we divided the features into two categories: spatial and temporal.
\subsubsection{Spatial}
Spatial features are demographic structures that select numeric columns from numeric columns, such as region/district, age, and race, and numeric counts from demographic prefixes, such as home\_*, male\_*, female\_*, and income\_*. Spatial features remain stable over time, which allows the model to capture long-term patterns. They are also called static features. 
\subsubsection{Temporal}
Temporal features mainly capture dynamic patterns such as hourly visits. A few temporal and contextual characteristics were created after cleaning to record periodic and behavioural dynamics of visitation, such as weekly rhythms and seasonal cycles. Those features are derived from the current date, such as the start of the week, the end of the week, the season, the month, and visitation a day before. This enables the model to learn short-term patterns.  
\begin{itemize}
\item Calendar information: weekend, month, and day of week (is weekend).

\item Seasonal grouping: winter (Dec-Feb), spring (Mar- May), summer (Jun- Aug), and fall (Sep-Nov).

\item Lag features: lag1 (number of visitors the day before) and lag7 (number of visitors the week before), a short-run change of direction.
\end{itemize}
To prevent data leakage and prevent biased learning, highly correlated features were removed. Some of the features are white, income\_60\_75, income\_75\_100, female\_45\_49, and so on. By eliminating these highly correlated features, we ensure predictive performance is meaningful and improve model robustness. The top 20  correlated features are described in the diagram. 
\begin{figure}[H]
    \centering
    \SafeIncludeGraphics[width=0.9\textwidth]{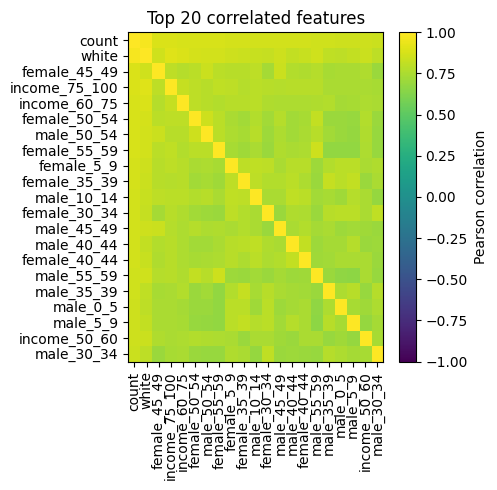}
    \caption{Heatmap for the top 20 highly correlated features}
    \label{fig:cor}
\end{figure}
\subsection{Remove Outliers}
\subsubsection{Anomaly Detection}
The anonymous dataset can contain duplicate signals, transmission errors, and irregular patterns. To identify this type of pattern, three anomaly detection methods are used. Different methods can give different results because there must be changes in high false positive results. To avoid that, multiple anomaly detection algorithms must be used to identify the true anomalies, which helps us to avoid relying on one algorithm. 

\begin{itemize}
\item Isolation Forest (IF) – isolates anomalous points by recursively partitioning the feature space.

\item Local Outlier Factor (LOF) – measures the local density deviation of a sample relative to its neighbours.

\item One-Class Support Vector Machine (OCSVM) – separates normal instances from anomalies by learning a tight boundary around the majority data.
\end{itemize}

Each algorithm produced an anomaly score. A consensus-based voting approach was applied, where any data point labelled as anomalous by at least two detectors was removed:
\[
A_i =
\begin{cases}
1, & \text{if at least two detectors mark sample } i \text{ as anomalous}, \\
0, & \text{otherwise}.
\end{cases}
\]
\subsubsection{Interquartile Range (IQR)}
\begin{figure}[H]
    \centering
    \SafeIncludeGraphics[width=0.9\textwidth]{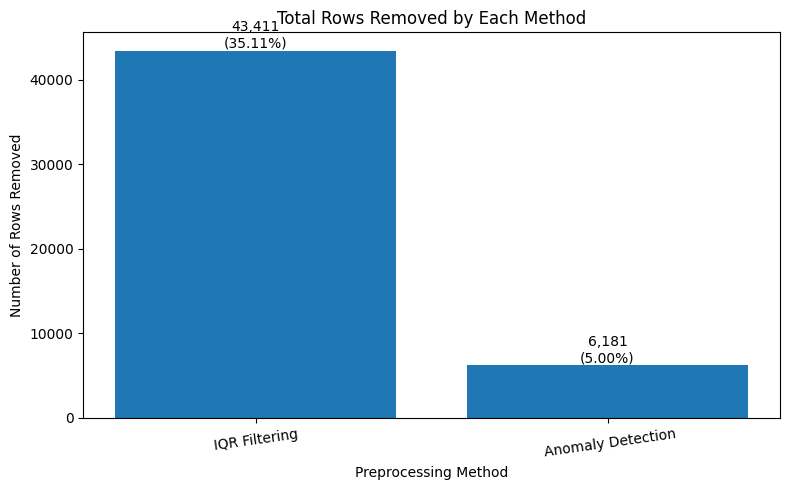}
    \caption{The number of rows that were removed from the data by the IQR filtering and anomaly detection during data pre-processing.}
    \label{fig:database}
\end{figure}

In order to identify and remove outliers, Interquartile Range (IQR) filtering was applied to the numeric predictor variables  before model training. This preprocessing step reduced the influence of extreme values and improved the quality of the dataset used for machine learning model development.
Figure \ref{fig:database} indicates the amount of data removed by the two techniques in preprocessing. Anomaly detection removed just 6,181 rows (5.00\%) of the data, whereas 43,411 rows (35.11\%) removed by IQR filtering. That indicates that the anomaly detection technique can preserve a large amount of data compared to IQR filtering.

\subsection{Base Model Candidates}
To evaluate performance in different condition we have selected different models. In this research, we considered three types of models: tree-based and ensemble models, linear and regularised models, and neural network models.
\subsubsection{Tree-based and Ensemble models}
Enable capturing non-linear patterns, we have used Random Forest (RF) and Extra Trees Regressor (ETR). RF helps to reduce variance. Besides, ETR improves generalisation. In addition, we have also employed a Gradient Boosting Regressor (GBR) to capture additive nonlinear relationships and a Decision Tree (DT) to provide hierarchical decision structures. 

\subsubsection{Linear and Regularized Models}
In order to consider linear baselines, we selected linear regression (LR) and ridge regression. Ridge regression used L2
to overcome overfitting. Moreover, Huber Regression and Bayesian Ridge Regression are also included to improve robustness and stability

\subsubsection{Neural Network Models}
We used more than one Multi-Layer Perceptron (MLP) architecture of different layers, such as (64), (128, 64), and (256, 128, 64) hidden units. Standardised inputs, ReLU activation, early stopping, and adaptive learning rates were also used to train all neural models so that they are stable in their convergence. They are very efficient at capturing high-dimensional nonlinear patterns from spatial and temporal features.

The three best-performing models were selected for subsequent ensemble evaluation.

\subsubsection{Evaluation Metrics}

We evaluate our proposed method using both regression metrics and classification metrics. Base models are so strong, and differences in prediction accuracy and loss can be very minimal, making it hard to identify the best models. Multiple metrics help to overcome that issue\cite{cite19}\cite{cite18}\cite{cite25}. Besides, the regression metrics are used to indicate the similarity of the predicted and observed visitation counts, whereas the classification metrics are used to define whether the model can be used to identify the most relevant, highest-demand periods that are important in the management decisions.

\paragraph{Mean Absolute Error (MAE).}
MAE examines the average prediction errors in terms of target visitation\cite{cite20}\cite{cite22}.  It is simple to interpret and use in nature as the average absolute deviation between mean predicted counts and counts.Lower values indicate that the model's performance is higher. 

\begin{equation}
\mathrm{MAE} = \frac{1}{n}\sum_{i=1}^{n}\left|y_i-\hat{y}_i\right|.
\end{equation}

where,
\begin{itemize}
    \item $n$ is the total number of observations.
    \item $y_i$ is the actual (observed) value.
    \item $\hat{y}_i$ is the predicted value.
\end{itemize}

\paragraph{Mean Squared Error (MSE)}
MSE computes the average squared residual, which indicates that a higher value degrades the model's performance\cite{cite20}. Moreover, MSE is more sensitive to rare extreme errors and may be dominated by a small number of large residuals. MSE is more susceptible to large-scale extreme errors and could be dominated by relatively few large residuals.

\begin{equation}
\mathrm{MSE} = \frac{1}{n}\sum_{i=1}^{n}\left(y_i-\hat{y}_i\right)^2.
\end{equation}

where,
\begin{itemize}
    \item $n$ is the total number of observations.
    \item $y_i$ is the actual (observed) value.
    \item $\hat{y}_i$ is the predicted value.
\end{itemize}

\paragraph{Coefficient of Determination ($R^2$).}
$R^2$ explains how correctly the model predict visitation.\cite{cite21}  
 Higher values indicate strong power. 
\begin{equation}
R^2 = 1 - \frac{\sum_{i=1}^{n}\left(y_i-\hat{y}_i\right)^2}{\sum_{i=1}^{n}\left(y_i-\bar{y}\right)^2},
\end{equation}
where $\bar{y}$ denotes the mean of the ground-truth values.

where,
\begin{itemize}
    \item $y_i$ is the actual (observed) value.
    \item $\hat{y}_i$ is the predicted value.
     \item $\bar{y}$ is the mean of the actual (observed) values.
\end{itemize}

\paragraph{Root Mean Square Error }
Root Mean Squared Error (RMSE) measures the average magnitude of the prediction error\cite{cite35}. It is the square root of the Mean Squared Error (MSE), giving the error in the same units as the target variable. The lower the RMSE the better the predictive performance. 

\begin{equation}
\mathrm{RMSE} =
\sqrt{
\frac{1}{n}
\sum_{i=1}^{n}
\left(y_i-\hat{y}_i\right)^2
}
\label{eq:rmse}
\end{equation}

where,
\begin{itemize}
    \item $n$ is the total number of observations.
    \item $y_i$ is the actual (observed) value.
    \item $\hat{y}_i$ is the predicted value.
\end{itemize}

\paragraph{Mean Absolute Percentage Error (MAPE)}

The Mean Absolute Percentage Error (MAPE) indicates the average prediction error as a percentage of the actual values\cite{cite36}. It is useful for comparing the accuracy of prediction within different  data sets at different scales. The smaller the value,  the more accurate the predictions.

\begin{equation}
\mathrm{MAPE} = \frac{100}{n}\sum_{i=1}^{n}
\left|
\frac{y_i-\hat{y}_i}{y_i}
\right|
\label{eq:mape}
\end{equation}

where,
\begin{itemize}
    \item $n$ is the total number of observations.
    \item $y_i$ is the actual (observed) value.
    \item $\hat{y}_i$ is the predicted value.
\end{itemize}

\subsection{Meta Model}
For creating an effective model, we constructed a hybrid metamodel using the top 3 base models that provide the highest performance. Using the same features and the same train-test split, we combined them using two ensemble strategies: (i) Voting regression to reduce variance, (ii) stacking regression to overcome systematic errors. In stacking, we used multiple meta-learners (e.g., Ridge and Gradient Boosting) and trained them using cross-validation. Hybrid models perform better because they integrate complementary predictors, making both models less biased and less variable and increasing the generalisation.

\section{Results}\label{sec2}
Table \ref{tab:iqr_lag_comparison} and  \ref{tab:anomaly_lag}  described the MAE, MSE, RMSE, MAPE, and R² of the best baseline models before applying the temporal feature named the lag feature. This scenario is captured before and after applying IQR and anomaly detection techniques. 

In term of MAE, both  Interquartile Range (IQR) filtering and anomaly removal showed improvement. The lag value indicates the past value of a particular feature. Lag features improved several models, although not all models benefited in every evaluation setting. For example, for the best base predictor before using IQR removal technique and considering the lag feature, the MLP (256,128,64) model decreased the test MAE from 1.705 to 1.323. Moreover, after applying IQR filtering for the same base model, the test R² value increased from 0.991 to 0.997 with a decrease in the test MAE value from 0.640 to 0.613, and a similar improvement was also observed in other metrics such as MSE, RMSE and MAPE. Likewise, after applying IQR and incorporating lag features, considering metrics like MSE, RMSE, and MAPE showed improvement for other predictors like  MLP(128,64) and ETR.  Moreover, some models, such as ETR, showed a substantially higher MSE before applying IQR filtering (2162.719), and the value dropped significantly to 3.011  that indicated that the IQR filtering method was quite effective in reducing the outliers and improved the predictive performance of some models, and all models performed well with the lag feature.

\begin{sidewaystable}[p]
\centering
\caption{Comparison of representative models before and after IQR filtering with and without lag features.}
\label{tab:iqr_lag_comparison}

\scriptsize
\setlength{\tabcolsep}{3.2pt}
\renewcommand{\arraystretch}{1.12}

\begin{tabular}{ll*{10}{c}}
\toprule

\multirow{2}{*}{\textbf{Stage}} &
\multirow{2}{*}{\textbf{Model}} &
\multicolumn{2}{c}{\textbf{MAE}} &
\multicolumn{2}{c}{\textbf{MSE}} &
\multicolumn{2}{c}{\textbf{RMSE}} &
\multicolumn{2}{c}{\textbf{MAPE (\%)}} &
\multicolumn{2}{c}{\textbf{$R^2$}} \\

\cmidrule(lr){3-4}
\cmidrule(lr){5-6}
\cmidrule(lr){7-8}
\cmidrule(lr){9-10}
\cmidrule(lr){11-12}

& &
\textbf{No Lag} &
\textbf{With Lag} &
\textbf{No Lag} &
\textbf{With Lag} &
\textbf{No Lag} &
\textbf{With Lag} &
\textbf{No Lag} &
\textbf{With Lag} &
\textbf{No Lag} &
\textbf{With Lag} \\

\midrule


\multirow{3}{*}{\shortstack{Before\\IQR\\Filtering}}

& \textbf{MLP (256,128,64)}
& 1.705 & \textbf{1.323}
& 11.738 & \textbf{5.715}
& 3.426 & \textbf{2.391}
& 5.001 & \textbf{4.818}
& \textbf{1.000} & \textbf{1.000} \\

& \textbf{MLP (128,64)}
& \textbf{1.404} & 1.426
& \textbf{6.458} & 6.833
& \textbf{2.541} & 2.614
& \textbf{4.545} & 4.805
& \textbf{1.000} & \textbf{1.000} \\

& \textbf{ETR}
& \textbf{4.468} & 4.746
& \textbf{1638.830} & 2162.719
& \textbf{40.482} & 46.505
& \textbf{1.732} & 1.828
& \textbf{0.994} & 0.992 \\

\midrule


\multirow{3}{*}{\shortstack{After\\IQR\\Filtering}}

& \textbf{MLP (256,128,64)}
& 0.640 & \textbf{0.613}
& 3.363 & \textbf{1.020}
& 1.834 & \textbf{1.010}
& 3.004 & \textbf{2.826}
& 0.991 & \textbf{0.997} \\

& \textbf{MLP (128,64)}
& 0.733 & \textbf{0.677}
& 4.018 & \textbf{1.259}
& 2.004 & \textbf{1.122}
& 3.548 & \textbf{3.232}
& 0.989 & \textbf{0.997} \\

& \textbf{ETR}
& 0.559 & \textbf{0.550}
& 7.428 & \textbf{3.011}
& 2.725 & \textbf{1.735}
& \textbf{1.823} & 1.859
& 0.980 & \textbf{0.992} \\

\bottomrule
\end{tabular}
\end{sidewaystable}


After anomaly detection, several models showed lower prediction errors, although improvements were not uniform across metrics or lag settings. Table~\ref{tab:anomaly_lag} compares MAE, MSE, RMSE, and MAPE before and after anomaly removal. With lag features after anomaly removal, Bayesian Ridge achieved the lowest MAE (0.880), while the MLP achieved the lowest MSE (2.292), RMSE (1.514), and MAPE (3.605\%). Moreover,  significant improvement was shown in MLP(256,128,64) as MAE dropped from 1.043 to 1.003, MSE reduced from 4.243 to 2.292, and other metrics like RMSE and RMSE also deceased through Anomaly removal and incorporating lag features. This result indicates that after using the anomaly identification approach, the lag feature improved performance. Table~\ref{tab:anomaly_lag} indicates that before using the anomaly process, all model were providing strong performance in terms of MAE and MSE WITH\_LAG conditions. Moreover, many models like Ridge, in terms of MAE, MSE,RMSE, and MAPE showed effective performance. This result reveals that the anomaly detection process enhanced performance for many models.

\begin{sidewaystable}[p]
\centering
\caption{Comparison of representative models before and after anomaly removal with and without the lag feature.}
\label{tab:anomaly_lag}

\scriptsize
\setlength{\tabcolsep}{3.2pt}
\renewcommand{\arraystretch}{1.12}

\begin{tabular}{ll*{10}{c}}
\toprule

\multirow{2}{*}{\textbf{Stage}} &
\multirow{2}{*}{\textbf{Model}} &
\multicolumn{2}{c}{\textbf{MAE}} &
\multicolumn{2}{c}{\textbf{MSE}} &
\multicolumn{2}{c}{\textbf{RMSE}} &
\multicolumn{2}{c}{\textbf{MAPE (\%)}} &
\multicolumn{2}{c}{\textbf{$R^2$}} \\

\cmidrule(lr){3-4}
\cmidrule(lr){5-6}
\cmidrule(lr){7-8}
\cmidrule(lr){9-10}
\cmidrule(lr){11-12}

&
&
\textbf{No Lag} &
\textbf{With Lag} &
\textbf{No Lag} &
\textbf{With Lag} &
\textbf{No Lag} &
\textbf{With Lag} &
\textbf{No Lag} &
\textbf{With Lag} &
\textbf{No Lag} &
\textbf{With Lag} \\

\midrule


\multirow{3}{*}{\shortstack{\textbf{Before}\\\textbf{Anomaly}\\\textbf{Removal}}}

& \textbf{Bayes}
& \textbf{0.890} & 0.891
& \textbf{2.949} & 4.494
& \textbf{1.717} & 2.120
& \textbf{3.574} & 3.617
& \textbf{1.000} & \textbf{1.000} \\

& \textbf{Ridge}
& \textbf{0.957} & 1.211
& \textbf{3.161} & 5.589
& \textbf{1.778} & 2.364
& \textbf{3.859} & 4.681
& \textbf{1.000} & \textbf{1.000} \\

& \textbf{MLP (256,128,64)}
& 1.600 & \textbf{1.437}
& 8.076 & \textbf{7.956}
& 2.842 & \textbf{2.821}
& \textbf{4.704} & 4.798
& \textbf{1.000} & \textbf{1.000} \\

\midrule


\multirow{3}{*}{\shortstack{\textbf{After}\\\textbf{Anomaly}\\\textbf{Removal}}}

& \textbf{Bayes}
& 0.889 & \textbf{0.880}
& 3.946 & \textbf{2.396}
& 1.987 & \textbf{1.548}
& 3.714 & \textbf{3.668}
& \textbf{1.000} & \textbf{1.000} \\

& \textbf{Ridge}
& 0.913 & \textbf{0.886}
& 4.002 & \textbf{2.411}
& 2.000 & \textbf{1.553}
& 3.815 & \textbf{3.682}
& \textbf{1.000} & \textbf{1.000} \\

& \textbf{MLP (256,128,64)}
& 1.043 & \textbf{1.003}
& 4.243 & \textbf{2.292}
& 2.060 & \textbf{1.514}
& 3.611 & \textbf{3.605}
& \textbf{1.000} & \textbf{1.000} \\

\bottomrule
\end{tabular}
\end{sidewaystable}

\begin{figure*}[!htbp]
    \centering

    \begin{subfigure}[t]{0.50\textwidth}
        \centering
        \SafeIncludeGraphics[width=\linewidth]{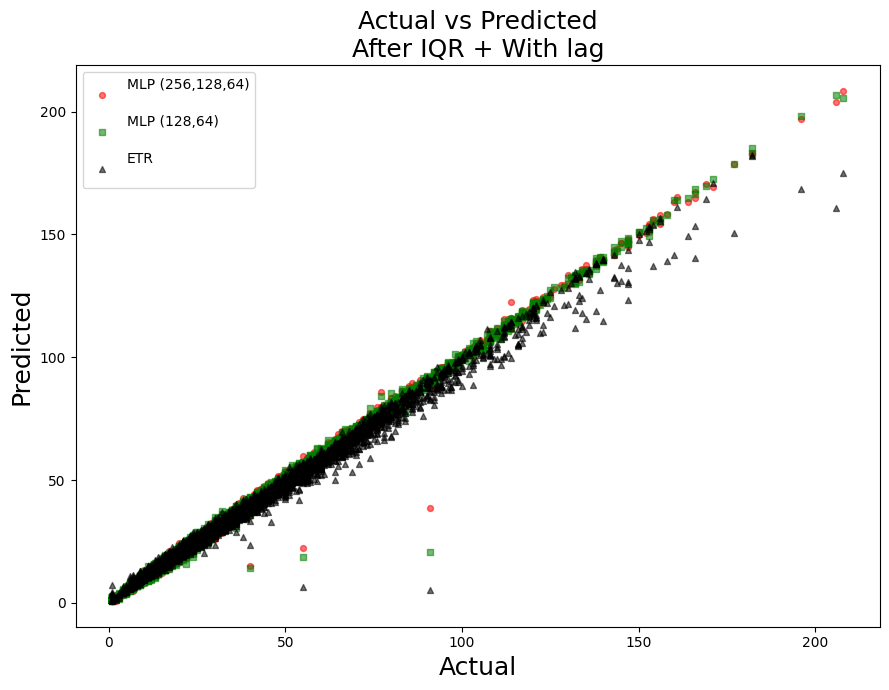}
        \caption{IQR Technique}
        \label{fig:base_iqr}
    \end{subfigure}\hfill
    \begin{subfigure}[t]{0.50\textwidth}
        \centering
        \SafeIncludeGraphics[width=\linewidth]{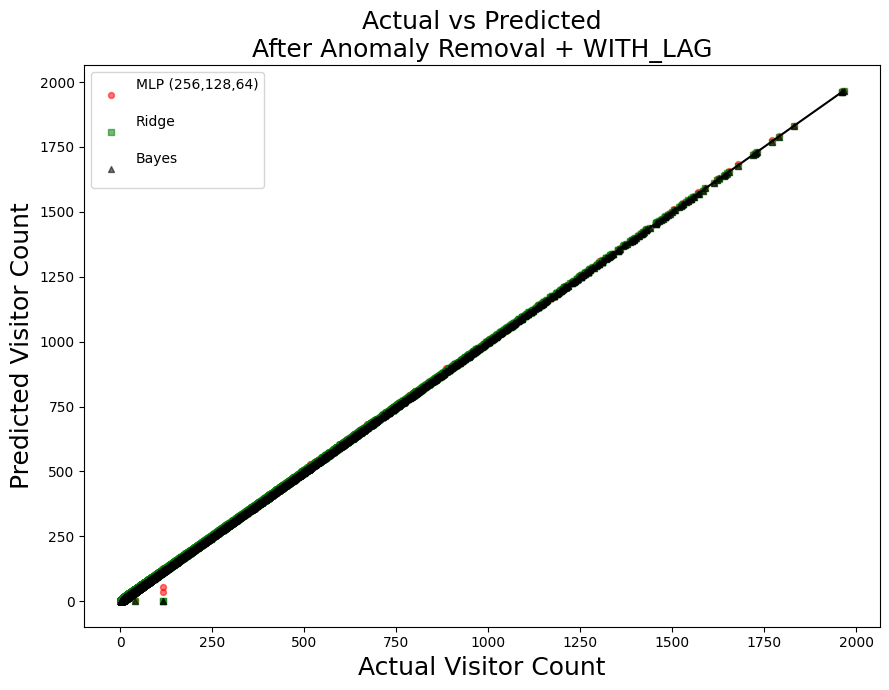}
        \caption{Anomaly Detection Technique}
        \label{fig:base_anomaly}
    \end{subfigure}

    \caption{Actual vs. predicted values for the base model after applying (a) the IQR technique and (b) the anomaly detection technique.}
    \label{fig:combined}
\end{figure*}

Figure~\ref{fig:combined} compares actual and predicted visitation for the base models, while Figure~\ref{fig:database} compares the number of observations removed by the two preprocessing methods.

\begin{figure*}[!htbp]
    \centering

    \begin{subfigure}[t]{0.50\textwidth}
        \centering
        \SafeIncludeGraphics[width=\linewidth]{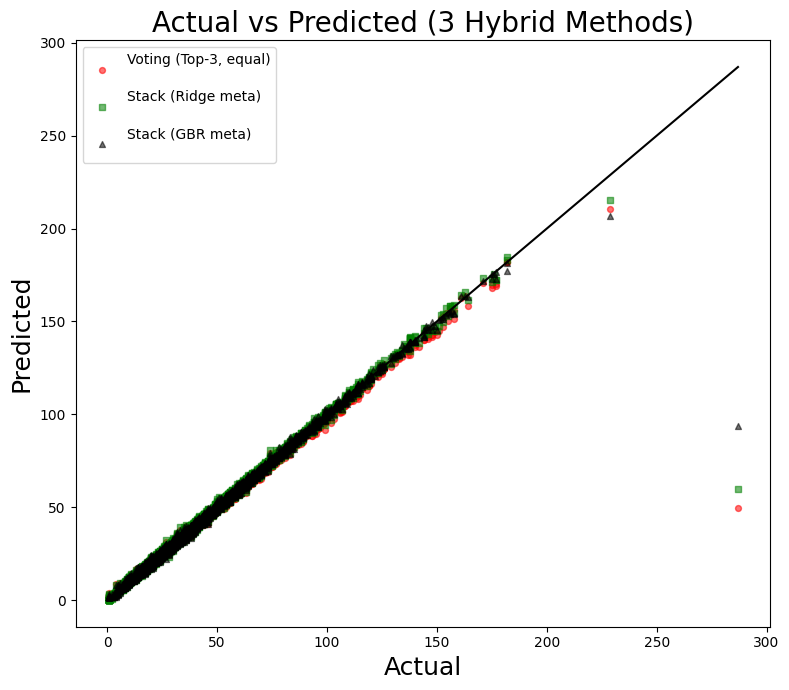}
        \caption{IQR Technique}
        \label{fig:hybrid_iqr}
    \end{subfigure}\hfill
    \begin{subfigure}[t]{0.50\textwidth}
        \centering
        \SafeIncludeGraphics[width=\linewidth]{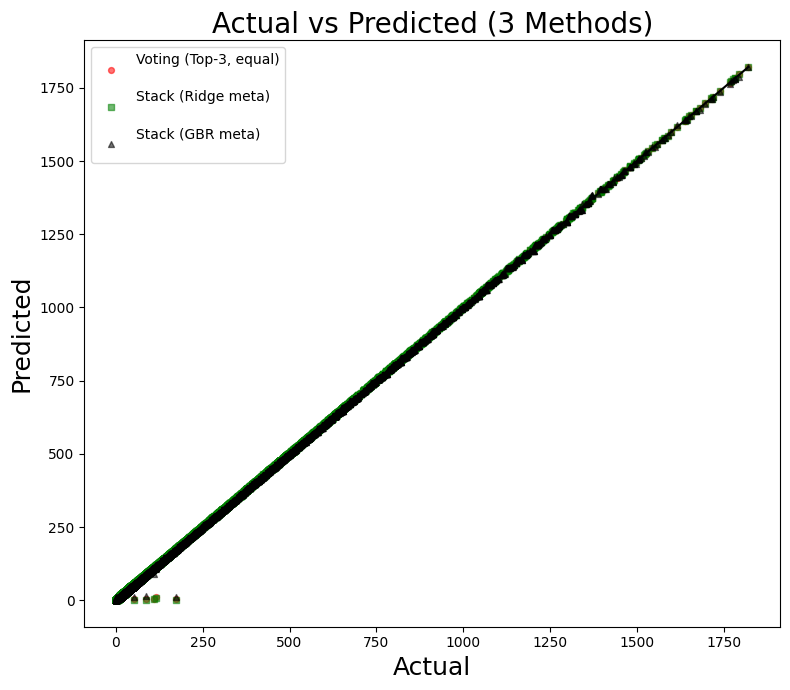}
        \caption{Anomaly Detection Technique}
        \label{fig:hybrid_anomaly}
    \end{subfigure}

    \caption{Actual vs. predicted values for the Hybrid after applying (a) the IQR technique and (b) the anomaly detection technique.}
    \label{fig:combined2}
\end{figure*}

Using the best three performing base models as inputs, the hybrid models were constructed and evaluated. Figures~\ref{fig:combined} and~\ref{fig:combined2}, and Table~\ref{tab:hybrid_results} showed that the impact of anomaly detection showed strong predictive performance with both base and meta-learners. Instead of low MAE and RMSE in base models, hybrid meta-models were more effective by achieving less error in terms of MAE and MAPE. For the single-split comparison, the voting model achieved an MAE of 0.862 and MAPE of 3.325\%; the lowest base-model MAE was 0.880, while the lowest base-model MAPE was 3.605\%. \ref{fig:combined} and \ref{fig:combined2} showed that, after applying anomaly detection, the difference between the predicted and the actual value for all models (base models and hybrid models) was very minimal, but \ref{tab:hybrid_results} showed that STACK\_GBR had an MAE of 0.951 and MAPE of 3.406\%; these results are not uniformly lower than those of the base models.

\begin{figure}[!htbp]
    \centering
     \SafeIncludeGraphics[height=0.50\textheight,keepaspectratio]{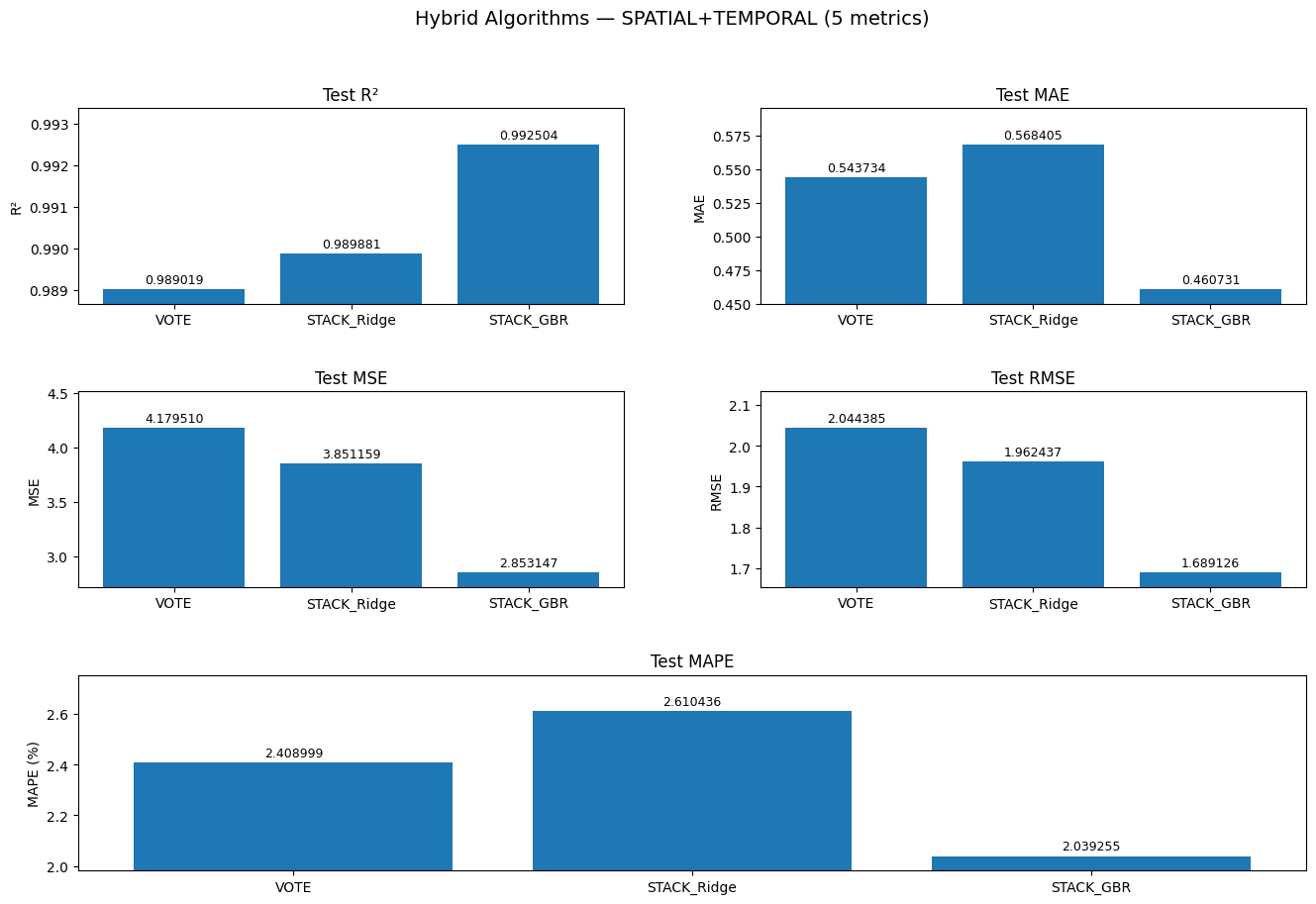}
    \caption{Anomaly -based evaluation of hybrid meta-models using R², MAE, MSE, RMSE, and MAPE.}
    \label{fig:comp3}
\end{figure}

\begin{figure}[!htbp]
    \centering
    \SafeIncludeGraphics[height=0.50\textheight]{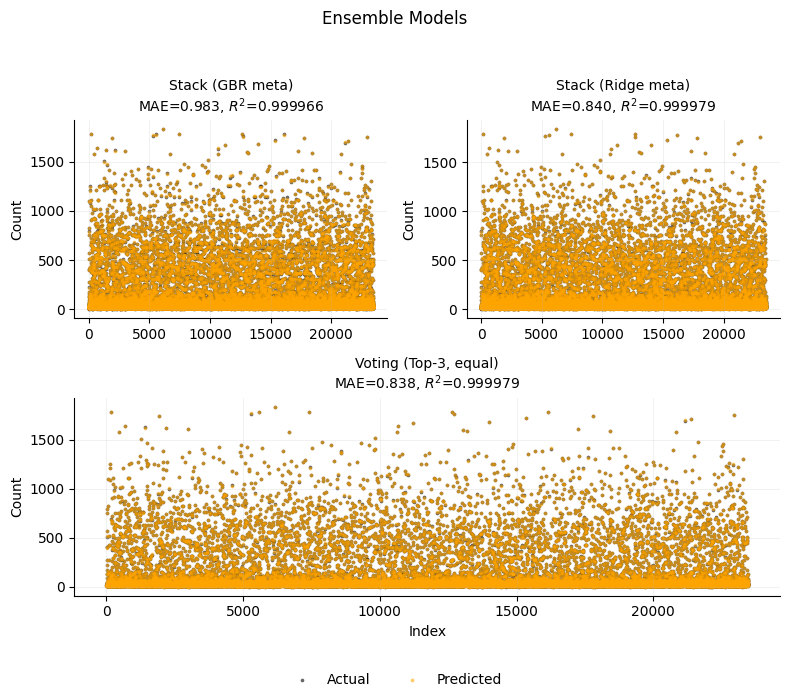}
    \caption{Scatter plot for actual vs. predicted values for voting, Ridge-based stacking, and GBR-based stacking}
    \label{fig:sac2}
\end{figure}

Table~\ref{tab:hybrid_results} and Fig.~\ref{fig:comp3} compared the predictive performance of the base models and the proposed hybrid models using five evaluation metrics: $R^2$, MAE, MSE, RMSE, and MAPE. Among the hybrid models in Table~\ref{tab:hybrid_results}, STACK\_GBR achieved the highest $R^2$ and lowest MSE and RMSE; VOTE achieved the lowest MAE, while STACK\_Ridge achieved the lowest MAPE. No single hybrid model led every metric. The other hybrid models were competitive on selected metrics, although not all outperformed the base models. These results indicate that the value of hybrid learning depends on the evaluation metric and setting. Moreover, Fig.~\ref{fig:sac2} presents the scatter plot for predicted vs actual result for hybrid meta-models.

\begin{table}[!htbp]
\centering
\caption{Performance comparison of the top three base models and top three hybrid models after anomaly removal with lag features.}
\label{tab:hybrid_results}
\begin{tabular}{llccccc}
\toprule
\textbf{Category} & \textbf{Model} & \textbf{MAE} & \textbf{MSE} & \textbf{$R^2$} & \textbf{RMSE} & \textbf{MAPE (\%)} \\
\midrule
\multirow{3}{*}{Base}
& MLP (256,128,64) & 1.003 & \textbf{2.292} & \textbf{1.0000} & \textbf{1.514} & \textbf{3.605} \\
& Ridge            & 0.886 & 2.411 & \textbf{1.0000} & 1.553 & 3.682 \\
& Bayesian Ridge   & \textbf{0.880} & 2.396 & \textbf{1.0000} & 1.548 & 3.668 \\
\midrule
\multirow{3}{*}{Hybrid}
& STACK\_GBR       & 0.951 & \textbf{3.099} & \textbf{0.999946} & \textbf{1.760} & 3.406 \\
& VOTE             & \textbf{0.862} & 3.687 & 0.999936 & 1.920 & 3.325 \\
& STACK\_Ridge     & 0.865 & 3.721 & 0.999935 & 1.929 & \textbf{3.310} \\
\bottomrule
\end{tabular}
\end{table}

\begin{figure}[!htbp]
    \centering
    \SafeIncludeGraphics[height=0.6\textheight, keepaspectratio]{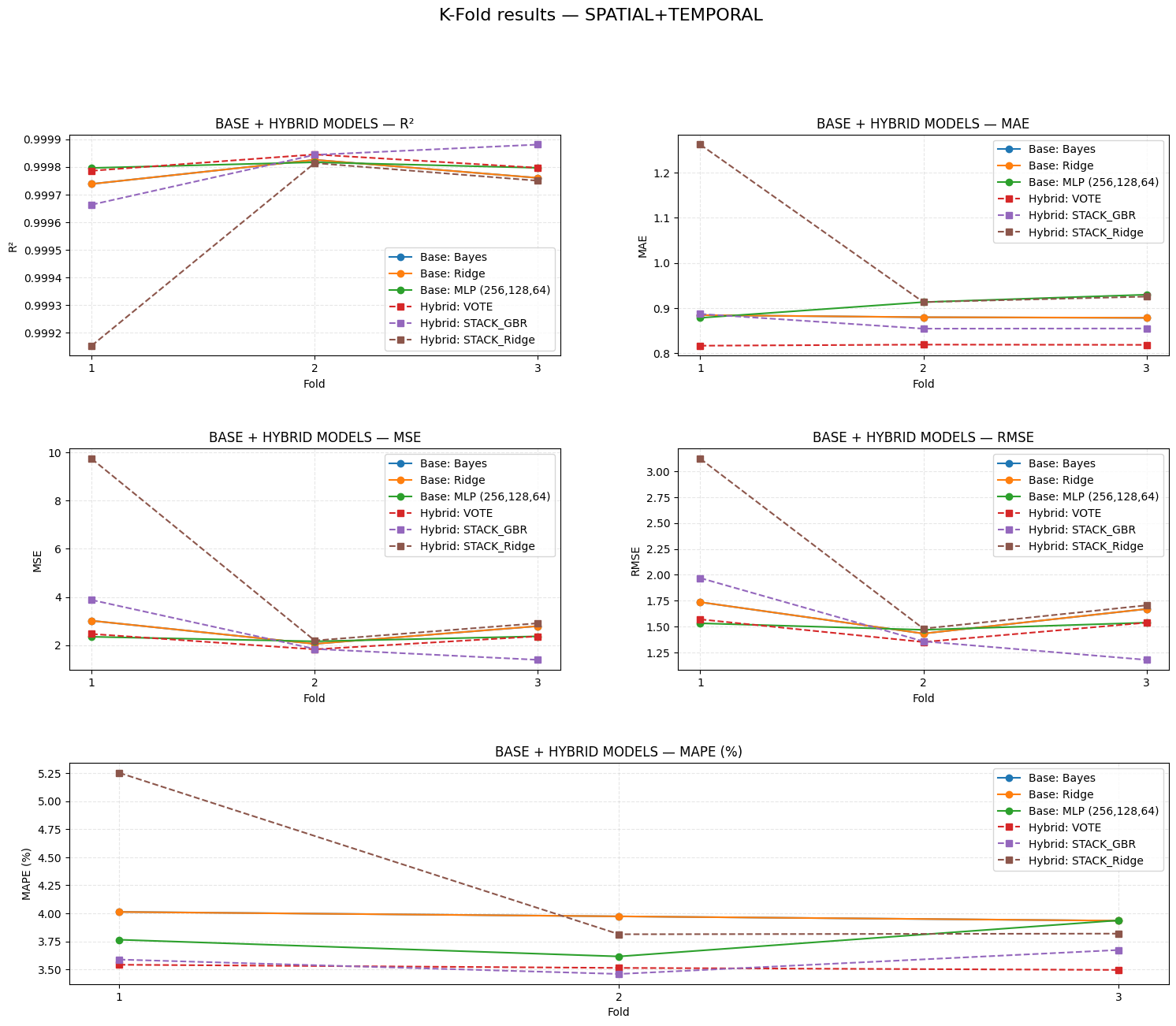}
    \caption{Cross-validation performance (MAE and R²) of base and hybrid models}
    \label{fig:kfold}
\end{figure}

\begin{figure}[!htbp]
    \centering
    \SafeIncludeGraphics[height=0.40\textheight, keepaspectratio]{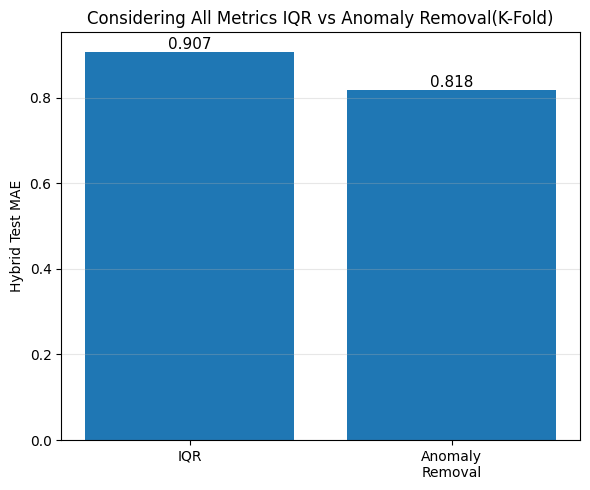}
    \caption{Using K-fold validation comparison of hybrid model test MAE using IQR filtering and anomaly removal}
    \label{fig:kfold2}
\end{figure}

\begin{figure}[!htbp]
    \centering
    \SafeIncludeGraphics[height=0.50\textheight]{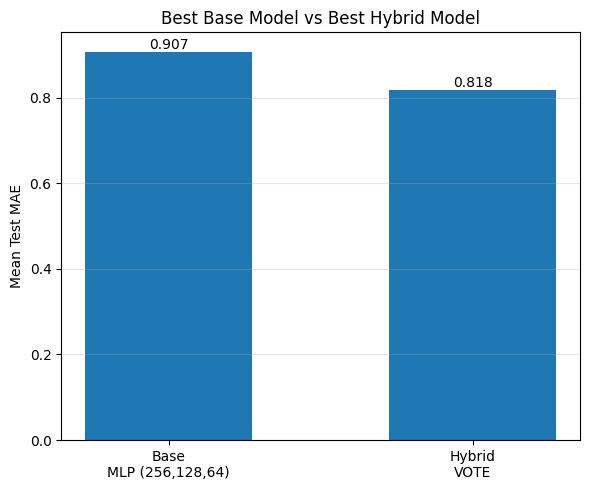}
    \caption{Base model vs Hybrid model using Test MAE }
    \label{fig:sac3}
\end{figure}

K-fold validation summary results are shown in Table \ref{tab:hybrid_iqr_kfold}. 
Different split results after applying anomaly detection techniques are shown in Figure~\ref{fig:kfold}indicating that the base model and hybrid model performances were similar and all models achieve higher accuracy ($R^2 > 99\%$). Table~\ref{tab:kfold_base_hybrid} reports aggregate model metrics rather than fold-level fluctuations. Conversely, ensemble approaches enhance stability: the voting model achieved the lowest MAE, MSE, RMSE, and MAPE and the highest $R^2$ in the anomaly-based cross-validation comparison. In the reported anomaly-based cross-validation comparison, VOTE achieved the best results on the evaluated metrics. Specifically, the hybrid model reduced MAE by 9.82\%, MSE by 3.12\%, RMSE by 1.76\%, and MAPE by 6.77\%, whereas STACK\_Ridge gave almost identical but higher error across folds, although with somewhat lower accuracy. Overall, the K-fold analysis favored VOTE among the evaluated models. From \ref{tab:hybrid_iqr_kfold} and ~\ref{tab:kfold_base_hybrid} depicted that, the best anomaly-based hybrid model (VOTE) reduced the MAE from 0.9070 to 0.8180 (9.81\% improvement), the MSE from 3.3776 to 2.2191 (34.30\% improvement), the RMSE from 1.8351 to 1.4866 (18.99\% improvement), and the MAPE from 3.5409\% to 3.5176\% (0.66\% improvement) compared with the best IQR-based hybrid model (STACK\_Ridge).Figure~\ref{fig:sac3} compares test MAE for the base and hybrid models after anomaly removal, illustrating the reported performance difference. Moreover, the MSE, RMSE, and MAPE of the hybrid model were also the lowest. When comparing the K-fold validation results of the IQR and anomaly detection techniques, Figure~\ref{fig:kfold2} shows that the anomaly detection technique performed better than IQR, with a 9.81\% improvement in MAE.

\begin{table}[!ht]
\centering
\caption{K-fold cross-validation performance of hybrid models after IQR-based outlier removal.}
\label{tab:hybrid_iqr_kfold}

\begin{tabular}{lccccc}
\hline
\textbf{Hybrid Models} & \textbf{MAE}$\downarrow$ & \textbf{MSE}$\downarrow$ & \textbf{RMSE}$\downarrow$ & \textbf{MAPE (\%)}$\downarrow$ & \textbf{R$^2$}$\uparrow$ \\
\hline
STACK\_Ridge & \textbf{0.906990} & \textbf{3.377647} & \textbf{1.835120} & 3.540943 & \textbf{0.999987} \\
VOTE         & 2.205798 & 300.798465 & 16.927260 & \textbf{3.392077} & 0.998871 \\
STACK\_GBR   & 4.531633 & 750.901946 & 23.888775 & 7.099465 & 0.997227 \\
\hline
\end{tabular}

\end{table}

\begin{table}[ht]
\centering
\caption{K-fold performance comparison of the base and hybrid models after anomaly removal}
\label{tab:kfold_base_hybrid}
\begin{tabular}{lccccc}
\toprule
\textbf{Model} &
\textbf{MAE} &
\textbf{MSE} &
\textbf{RMSE} &
\textbf{MAPE (\%)} &
\textbf{$R^2$} \\
\midrule

\multicolumn{6}{c}{\textbf{Base Models}}\\
\midrule
MLP (256,128,64) & 0.907074 & \textbf{2.290680} & \textbf{1.513177} & \textbf{3.773141} & \textbf{0.999803}\\
Bayes            & \textbf{0.880754} & 2.618634 & 1.612985 & 3.973741 & 0.999774\\
Ridge            & 0.880911 & 2.619215 & 1.613165 & 3.973542 & 0.999774\\

\midrule
\multicolumn{6}{c}{\textbf{Hybrid Models}}\\
\midrule
VOTE             & \textbf{0.818019} & \textbf{2.219127} & \textbf{1.486550} & \textbf{3.517616} & \textbf{0.999809}\\
STACK\_GBR       & 0.865331 & 2.370002 & 1.501891 & 3.574142 & 0.999796\\
STACK\_Ridge     & 1.033668 & 4.946523 & 2.102236 & 4.295037 & 0.999572\\

\bottomrule
\end{tabular}
\end{table}

\begin{figure}[!htbp]
    \centering

    \begin{subfigure}{0.49\textwidth}
        \centering
        \SafeIncludeGraphics[width=\linewidth]{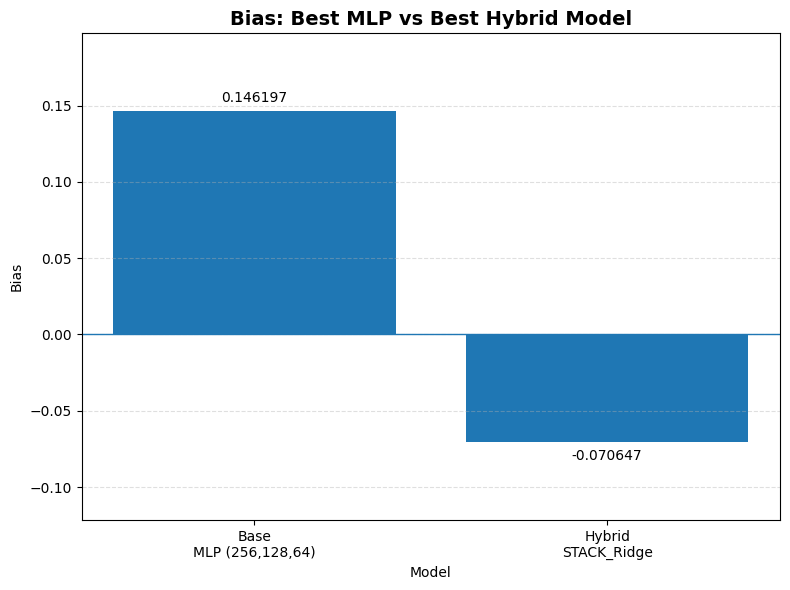}
        \caption{IQR Bias}
        \label{fig:bias_iqr}
    \end{subfigure}
    \hfill
    \begin{subfigure}{0.49\textwidth}
        \centering
        \SafeIncludeGraphics[width=\linewidth]{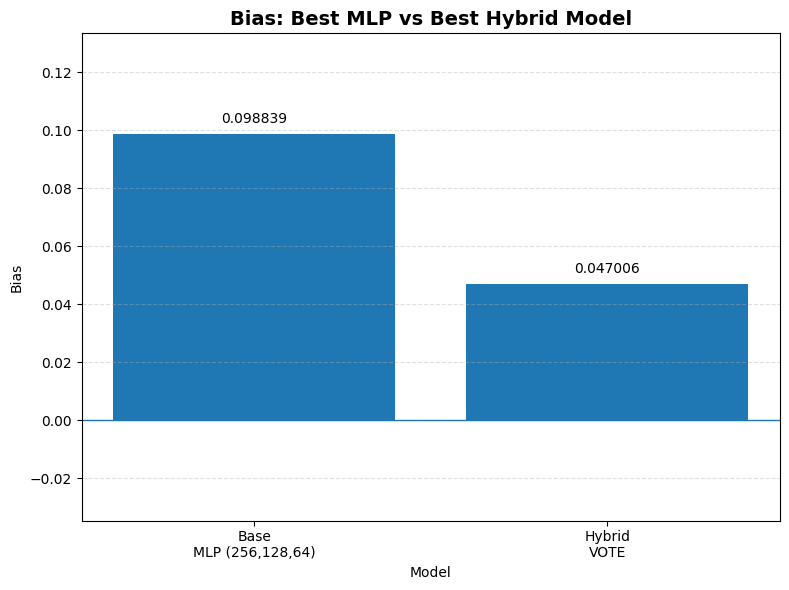}
        \caption{Anomaly Bias}
        \label{fig:bias_anomaly}
    \end{subfigure}

    \caption{Bias comparison of single and hybrid models using (a) IQR and (b) anomaly detection techniques.}
    \label{fig:bias}
\end{figure}

\begin{figure}[!htbp]
    \centering

    \begin{subfigure}{0.49\textwidth}
        \centering
        \SafeIncludeGraphics[width=\linewidth]{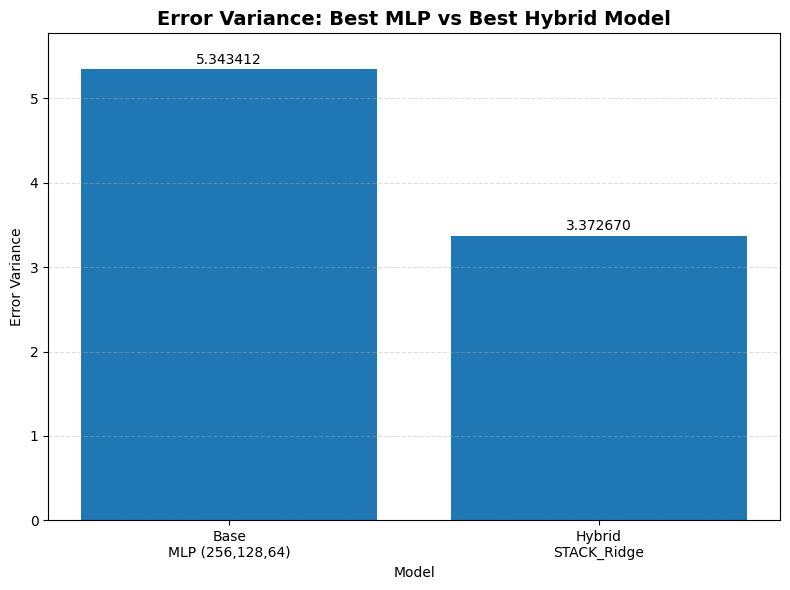}
        \caption{IQR Variance}
        \label{fig:var_iqr}
    \end{subfigure}
    \hfill
    \begin{subfigure}{0.49\textwidth}
        \centering
        \SafeIncludeGraphics[width=\linewidth]{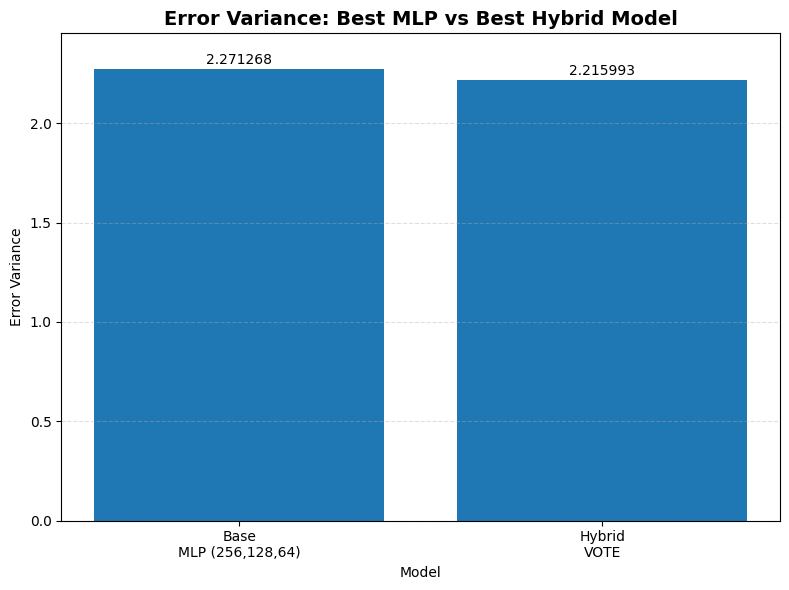}
        \caption{Anomaly Variance}
        \label{fig:var_anomaly}
    \end{subfigure}

    \caption{Prediction variance of single and hybrid models using (a) IQR and (b) anomaly detection techniques.}
    \label{fig:var}
\end{figure}

\begin{figure}[!htbp]
    \centering
    \SafeIncludeGraphics[height=0.22\textheight, keepaspectratio]{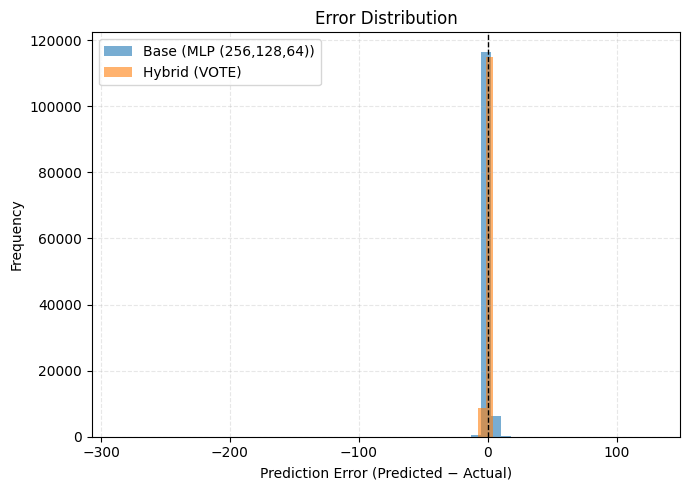}
    \caption{Error distribution of single vs. hybrid models}
    \label{fig:both}
\end{figure}

When bias is close to zero, it is good for the model, and when it is negative or positive, bias indicates underprediction or overprediction. Diagram \ref{fig:bias}, and \ref{fig:var} depicted that, considering IQR, for the hybrid model, negative bias was showing, and variance was showing less than the base model. When compared the variance with the anomaly identification technique was much greater; this indicated that after applying the anomaly detection technique, the model could capture more diverse prediction patterns. 
 After applying the anomaly identification technique, the hybrid model is more efficient, as it reduced systematic error (bias) while maintaining the same error variance. . Diagram \ref{fig:both} showed that both models are centred to zero, indicating less prediction error. Moreover, the hybrid model showed a narrower distribution, suggesting less variability and more consistent predictions across samples, thus enhancing overall predictive reliability. 

\section{Discussion}
\subsection*{Conclusions}
This study investigated the effectiveness of combining spatial and temporal features and lag features with ensemble learning to improve visitation count prediction accuracy. Experimental results showed that lag features and anomaly removal reduced prediction error in several settings, with high reported $R^2$ values for many models. Among individual learners, Bayes,Ridge,and multilayer perceptrons achieved consistently high performance. Moreover, hybrid approaches further enhanced robustness and error stability. Anomaly detection removed fewer records than IQR filtering and produced lower errors for the best hybrid models in the reported cross-validation comparison; the differing retained datasets should be considered when interpreting this result. 

The hybrid methods showed different strengths across evaluation metrics. In the reported anomaly-based cross-validation results, VOTE achieved the lowest MAE, MSE, RMSE, and MAPE and the highest $R^2$ among the compared models. These findings support the usefulness of ensemble methods in this evaluation, without establishing that every hybrid model outperformed every base model. The reported cross-validation summary favored VOTE on the evaluated metrics; however, the table does not provide fold-by-fold variability, so stability across folds cannot be established from those summary values alone. The k-fold validation experiment investigated that the VOTE method worked effectively.  Moreover, bias–variance analysis revealed that the proposed hybrid framework substantially decreases prediction bias without increasing variance, indicating improved reliability for real-world deployment.

Overall, the results confirm that integrating anomaly detection, temporal lag features, and hybrid learning provides a robust and accurate framework for large-scale spatiotemporal visitation modelling. Future work may extend this framework by incorporating extra features such as weather, adaptive ensemble weighting, and uncertainty quantification to further enhance operational decision support for resource planning and environmental management.




\IfFileExists{sn-bibliography.bib}{%
  \bibliographystyle{unsrtnat}%
  \bibliography{sn-bibliography}%

@inproceedings{cite1,
author = {Yap, Norman and Gong, Mingwei and Naha, Ranesh and Mahanti, Aniket},
year = {2020},
month = {10},
pages = {1-7},
title = {Machine Learning-based Modelling for Museum Visitations Prediction},
doi = {10.1109/ISNCC49221.2020.9297182},
booktitle = {Proceedings of the Conference}
}

@article{cite2,
author = {Zainoren, Abang and Wan Yaacob, Wan Fairos and Md Nasir, Syerina Azlin and Jaya, Serah and Mokhtar, Suhaili},
year = {2022},
month = {02},
pages = {1-16},
title = {Using Machine Learning to Predict Visitors to Totally Protected Areas in Sarawak, Malaysia},
volume = {14},
journal = {Sustainability},
doi = {10.3390/su14052735}
}

@article{cite3,
author = {Zainoren, Abang and Wan Yaacob, Wan Fairos and Md Nasir, Syerina Azlin and Jaya, Serah and Mokhtar, Suhaili},
year = {2022},
month = {02},
pages = {1-16},
title = {Using Machine Learning to Predict Visitors to Totally Protected Areas in Sarawak, Malaysia},
volume = {14},
journal = {Sustainability},
doi = {10.3390/su14052735}
}

@article{cite4,
author = {Sancho Núñez, José Carlos and Gómez-Pulido, Juan A. and Ramírez, Rafael},
year = {2024},
month = {07},
pages = {},
title = {Machine learning applied to tourism: A systematic review},
volume = {14},
journal = {Wiley Interdisciplinary Reviews: Data Mining and Knowledge Discovery},
doi = {10.1002/widm.1549}
}

@misc{cite5,
author = {Lemmel, Julian and Babaiee, Zahra and Kleinlehner, Marvin and Majic, Ivan and Neubauer, Philipp and Scholz, Johannes and Grosu, Radu and Neubauer, Sophie},
year = {2023},
month = {08},
pages = {},
title = {Prediction of Tourism Flow with Sparse Geolocation Data},
doi = {10.48550/arXiv.2308.14516}
}

@Article{cite6,
AUTHOR = {Louati, Ali and Louati, Hassen and Alharbi, Meshal and Kariri, Elham and Khawaji, Turki and Almubaddil, Yasser and Aldwsary, Sultan},
TITLE = {Machine Learning and Artificial Intelligence for a Sustainable Tourism: A Case Study on Saudi Arabia},
JOURNAL = {Information},
VOLUME = {15},
YEAR = {2024},
NUMBER = {9},
ARTICLE-NUMBER = {516},
URL = {https://www.mdpi.com/2078-2489/15/9/516},
ISSN = {2078-2489},
DOI = {10.3390/info15090516}
}

@inproceedings{cite7,
  title={iBAT: detecting anomalous taxi trajectories from GPS traces},
  author={Zhang, Daqing and Li, Nan and Zhou, Zhi-Hua and Chen, Chao and Sun, Lin and Li, Shijian},
  booktitle={Proceedings of the 13th international conference on Ubiquitous computing},
  pages={99--108},
  year={2011}
}

@article{cite8,
  title={Spatio-temporal analytics for exploring human mobility patterns and urban dynamics in the mobile age},
  author={Gao, Song},
  journal={Spatial Cognition \& Computation},
  volume={15},
  number={2},
  pages={86--114},
  year={2015},
  publisher={Taylor \& Francis}
}

@misc{cite9,
author = {Hamdi, Ali and Shaban, Khaled and Erradi, Abdelkarim and Mohamed, Amr and Rumi, Shakila and Shao, Wei},
year = {2021},
month = {03},
pages = {},
title = {Spatiotemporal Data Mining: A Survey on Challenges and Open Problems},
doi = {10.48550/arXiv.2103.17128}
}

@article{cite10,
author = {Cheng, Ximeng and Wang, Zhiqian and Yang, Xuexi and Xu, Liyan and Liu, Yu},
year = {2021},
month = {07},
pages = {101627},
title = {Multi-scale detection and interpretation of spatio-temporal anomalies of human activities represented by time-series},
volume = {88},
journal = {Computers, Environment and Urban Systems},
doi = {10.1016/j.compenvurbsys.2021.101627}
}

@article{cite11,
  title={Anomaly detection for volunteered geographic information: a case study of Safecast data},
  author={Xin, Yanan},
  journal={International Journal of Geographical Information Science},
  volume={36},
  number={7},
  pages={1423--1442},
  year={2022},
  publisher={Taylor \& Francis}
}

@article{cite12,
  title={Hyperparameters and tuning strategies for random forest},
  author={Probst, Philipp and Wright, Marvin N and Boulesteix, Anne-Laure},
  journal={Wiley Interdisciplinary Reviews: data mining and knowledge discovery},
  volume={9},
  number={3},
  pages={e1301},
  year={2019},
  publisher={Wiley Online Library}
}

@article{cite13,
  title={Gradient boosting machines, a tutorial},
  author={Natekin, Alexey and Knoll, Alois},
  journal={Frontiers in neurorobotics},
  volume={7},
  pages={21},
  year={2013},
  publisher={Frontiers Media SA}
}

@article{cite14,
  title={Forecasting tourist arrivals using STL-XGBoost method},
  author={He, Minmin and Qian, Xiyuan},
  journal={Tourism Economics},
  pages={13548166241313411},
  year={2025},
  publisher={SAGE Publications Sage UK: London, England}
}

@article{cite15,
  title={Anomaly detection based on isolation mechanisms: A survey},
  author={Cao, Yang and Xiang, Haolong and Zhang, Hang and Zhu, Ye and Ting, Kai Ming},
  journal={Machine Intelligence Research},
  volume={22},
  number={5},
  pages={849--865},
  year={2025},
  publisher={Springer}
}

@article{cite16,
  title={On the evaluation of unsupervised outlier detection: measures, datasets, and an empirical study},
  author={Campos, Guilherme O and Zimek, Arthur and Sander, J{\"o}rg and Campello, Ricardo JGB and Micenkov{\'a}, Barbora and Schubert, Erich and Assent, Ira and Houle, Michael E},
  journal={Data mining and knowledge discovery},
  volume={30},
  number={4},
  pages={891--927},
  year={2016},
  publisher={Springer}
}

@article{cite17,
  title={A review of novelty detection},
  author={Pimentel, Marco AF and Clifton, David A and Clifton, Lei and Tarassenko, Lionel},
  journal={Signal processing},
  volume={99},
  pages={215--249},
  year={2014},
  publisher={Elsevier}
}

@article{cite18,
  title={Evaluation metrics and statistical tests for machine learning},
  author={Rainio, Oona and Teuho, Jarmo and Kl{\'e}n, Riku},
  journal={Scientific Reports},
  volume={14},
  number={1},
  pages={6086},
  year={2024},
  publisher={Nature Publishing Group UK London}
}

@article{cite19,
  title={A critical review on selecting performance evaluation metrics for supervised machine learning models in wastewater quality prediction},
  author={Khoshvaght, Hoda and Permala, Ratish Ramyad and Razmjou, Amir and Khiadani, Mehdi},
  journal={Journal of Environmental Chemical Engineering},
  volume={13},
  number={6},
  pages={119675},
  year={2025},
  publisher={Elsevier}
}

@article{cite20,
  title={Comparative evaluation and comprehensive analysis of machine learning models for regression problems},
  author={Sekeroglu, Boran and Ever, Yoney Kirsal and Dimililer, Kamil and Al-Turjman, Fadi},
  journal={Data Intelligence},
  volume={4},
  number={3},
  pages={620--652},
  year={2022},
  publisher={MIT Press One Rogers Street, Cambridge, MA 02142-1209, USA journals-info~…}
}

@article{cite21,
  title={The coefficient of determination R-squared is more informative than SMAPE, MAE, MAPE, MSE and RMSE in regression analysis evaluation},
  author={Chicco, Davide and Warrens, Matthijs J and Jurman, Giuseppe},
  journal={Peerj computer science},
  volume={7},
  pages={e623},
  year={2021},
  publisher={PeerJ Inc.}
}

@article{cite22,
  title={Root mean square error (RMSE) or mean absolute error (MAE)?--Arguments against avoiding RMSE in the literature},
  author={Chai, Tianfeng and Draxler, Roland R},
  journal={Geoscientific model development},
  volume={7},
  number={3},
  pages={1247--1250},
  year={2014},
  publisher={Copernicus Publications G{\"o}ttingen, Germany}
}

@article{cite25,
  title={A survey of forecast error measures},
  author={Shcherbakov, Maxim Vladimirovich and Brebels, Adriaan and Shcherbakova, Nataliya Lvovna and Tyukov, Anton Pavlovich and Janovsky, Timur Alexandrovich and Kamaev, Valeriy Anatol’evich and others},
  journal={World Applied Sciences Journal},
  volume={24},
  number={24},
  pages={171--176},
  year={2013}
}

@article{cite27,
  title= {The use of tracking technologies in tourism research :The first decade},
  author={Shoval, Noam and Ahas, Rein},
  journal={Tourism Geographies},
  volume={18},
  number={5},
  pages={587--606},
  year={2016},
  publisher={Taylor \& Francis}
}

@article{cite28,
  title={A survey of results on mobile phone datasets analysis},
  author={Blondel, Vincent D and Decuyper, Adeline and Krings, Gautier},
  journal={EPJ data science},
  volume={4},
  number={1},
  pages={10},
  year={2015},
  publisher={Springer}
}

@article{cite30,
  title={LSTM-based deep learning models for long-term tourism demand forecasting},
  author={Salamanis, Athanasios and Xanthopoulou, Georgia and Kehagias, Dionysios and Tzovaras, Dimitrios},
  journal={Electronics},
  volume={11},
  number={22},
  pages={3681},
  year={2022},
  publisher={MDPI}
}

@article{cite31,
  title={Tourism demand forecasting: a deep learning model based on spatial-temporal transformer},
  author={Chen, Jiaying and Li, Cheng and Huang, Liyao and Zheng, Weimin},
  journal={Tourism Review},
  volume={80},
  number={3},
  pages={648--663},
  year={2025},
  publisher={Emerald Publishing Limited}
}

@article{cite33,
  title={Tracking tourists’ travel with smartphone-based GPS technology: a methodological discussion},
  author={Hardy, Anne and Hyslop, Sarah and Booth, Kate and Robards, Brady and Aryal, Jagannath and Gretzel, Ulrike and Eccleston, Richard},
  journal={Information Technology \& Tourism},
  volume={17},
  number={3},
  pages={255--274},
  year={2017},
  publisher={Springer}
}

@article{cite34,
  title={Understanding the bias of mobile location data across spatial scales and over time: a comprehensive analysis of SafeGraph data in the United States},
  author={Li, Zhenlong and Ning, Huan and Jing, Fengrui and Lessani, M Naser},
  journal={Plos one},
  volume={19},
  number={1},
  pages={e0294430},
  year={2024},
  publisher={Public Library of Science San Francisco, CA USA}
}

@article{cite35,
  title={Root mean square error (RMSE) or mean absolute error (MAE): When to use them or not},
  author={Hodson, Timothy O},
  journal={Geoscientific Model Development Discussions},
  volume={2022},
  pages={1--10},
  year={2022},
  publisher={G{\"o}ttingen, Germany}
}

@article{cite36,
  title={Enhancing museum visitor forecasting using deep learning and sentiment analysis: A transformer-based approach for sustainable management},
  author={Tian, Ziyi and Wang, Xiao and Wang, Yan and Lee, Jae Ho},
  journal={Plos one},
  volume={20},
  number={11},
  pages={e0335623},
  year={2025},
  publisher={Public Library of Science San Francisco, CA USA}
}
}{%
  \section*{References}
  \noindent\emph{Bibliography file sn-bibliography.bib has not been supplied. Add it to the Overleaf project to generate the reference list.}%
}

\end{document}